\documentclass[letterpaper, 10 pt, conference]{ieeeconf}  %

\IEEEoverridecommandlockouts                              %

\usepackage{graphics} %
\usepackage{epsfig} %
\usepackage{mathptmx} %
\usepackage{times} %
\usepackage{amsmath} %
\usepackage{amssymb}  %
\usepackage[table]{xcolor} %
\usepackage{hyperref}
\usepackage{mathtools}

\definecolor{plasmaPurple}{HTML}{0D0887} %
\definecolor{plasmaMagenta}{HTML}{9C179E}
\definecolor{plasmaOrange}{HTML}{ED6925}
\definecolor{plasmaYellow}{HTML}{F0F921} %

\title{\LARGE \bf
A Narwhal-Inspired Sensing-to-Control Framework\\ for Small Fixed-Wing Aircraft
}

\author{Fengze Xie$^{*1}$, Xiaozhou Fan$^{*1}$, Jacob Schuster$^{1}$, Yisong Yue$^{1}$, Morteza Gharib$^{1}$%
\thanks{This work is supported by the Technology Innovation Institute. $^{*}$ Equal contribution. $^{1}$ California Institute of Technology, CA 91106, USA. Corresponding author: Xiaozhou Fan (xiaozhoufan@hkust-gz.edu.cn), current position: Assistant Professor at HKUST (Guangzhou).\tt\small}
}

\begin{document}
\maketitle

\setcounter{page}{1}
\thispagestyle{empty}
\pagestyle{empty}

\begin{abstract}
Fixed-wing unmanned aerial vehicles (UAVs) offer endurance and efficiency but lack low-speed agility because its highly-coupled dynamical model. We present an end-to-end sensing-to-control pipeline that combines bio-inspired hardware instrumentation, physics-informed dynamics learning, and convex control allocation. Measuring airflow on a small airframe is difficult as near-body aerodynamics, propeller slipstream, control surfaces actuation, and the present of gusts would distort pressure signals. Inspired by the narwhal whale's signature protruding tusk, we stick our in-house developed multi-hole probes far into the upstream, and complement it with sparse yet carefully placed wing pressure sensors for local flow measurement. A data-driven calibration scheme was adopted to map the pressure signal of the probes to airspeed and flow angles. We next learn a control-affine dynamical model using Pitot-tube estimated airspeed and flow angles as inputs, along with sparse sensor measurements. We implement a soft left/right symmetry regularizer that improves the model's identifiability under partial observability and limits confounding between wing pressures and flaperon inputs. Desired wrenches (forces and moments output) are realized by a regularized least-squares optimizer that yields smooth, trimmed actuation. Wind tunnel studies, across a wide range of parameter space, demonstrate that adding wing pressures reduces force-estimation error by \(25\%\)–\(30\%\), the proposed model degrades less under distribution shift (about \(12\%\) versus \(44\%\) for an unstructured baseline), and force tracking improves with smoother inputs, including a \(27\%\) reduction in normal-force RMSE relative to a plain affine model and \(34\%\) relative to an unstructured baseline.
\end{abstract}

\section{Introduction}

\begin{figure}[t]
    \centering
    \includegraphics[width=0.9\linewidth]{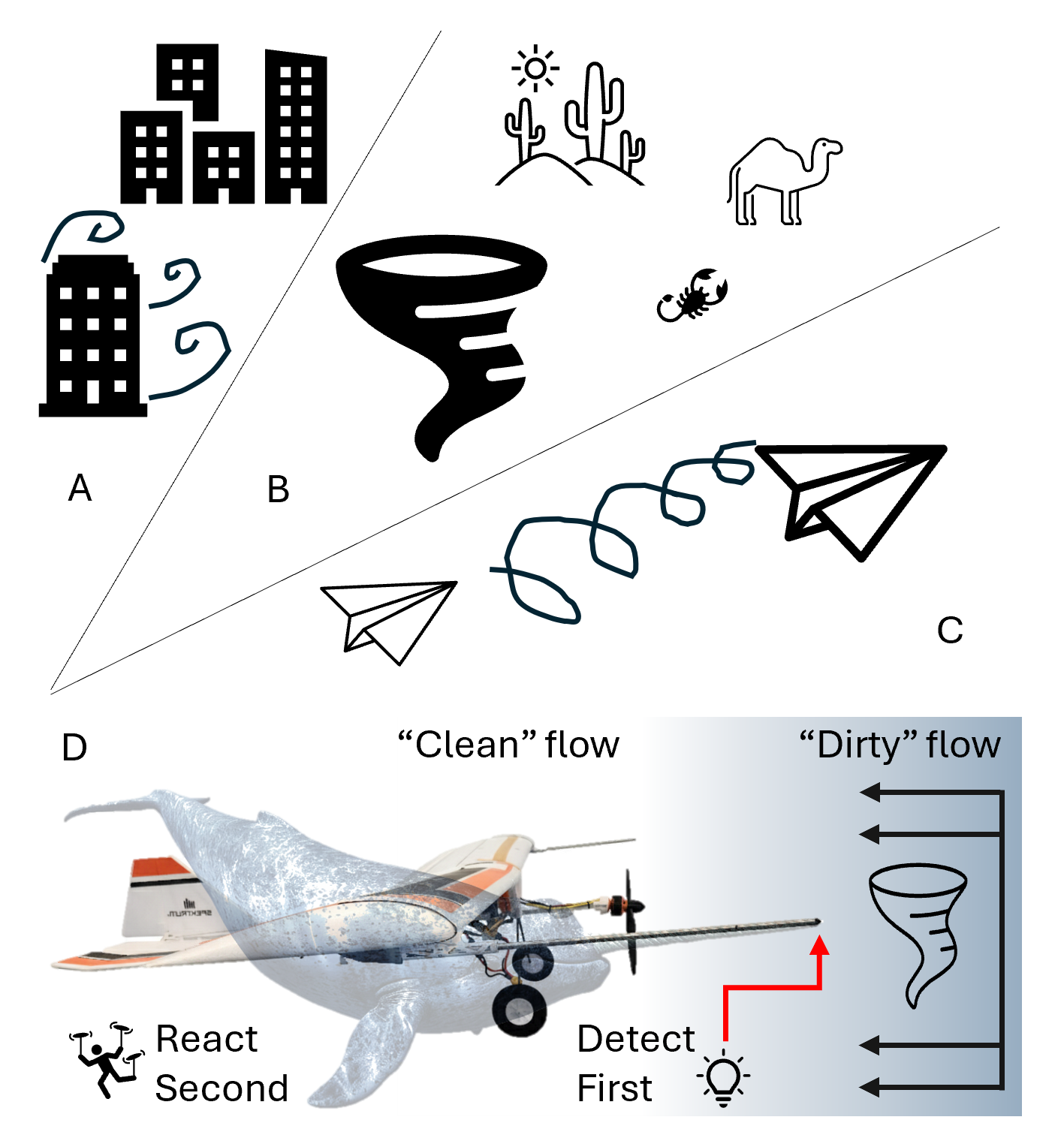}
    \caption{Fixed-wing vehicles operate across highly diverse terrain and environments, such as (A) through populated urban environments, (B) across vast deserts, or (C) trailing behind other aerial vehicles in formation flight. The underlying challenging gust are vortex dynamics which are highly time-varying. (D) To stay ahead of the ``curve'', we built in-house developed, avionic sensors that measures incoming gust (in loose language, "dirty" flow) before it arrives at the wing, and affect drone's dynamics. This approach is inspired by the Narwhal whale, where its tooth protrudes forward, and detects sea water salinity and potentially pressure signals \cite{nweeia2014}.}
    \label{fig:intro}
\end{figure}

Unmanned aerial vehicles (UAVs) have progressed rapidly across multiple platform classes. There are three major categories: flapping-wing, quadrotor, and fixed-wing aerial vehicles. Hybrid models are common. For example, a VTOL platform can use rotorcraft lift for takeoff and landing and a fixed-wing layout for efficient cruise~\cite{Shi2020}. Flapping wing vehicles are superior at birds- or insect-size due to their capability to harvest unsteadiness in the flow \cite{shyy2013, windes2018}, and has seen considerable research in understanding its mechanics with robotics instrumentation \cite{fan2024,fan2025}, but may still be years from becoming viable competitors to fixed-wing or quadrotors, due to their extremely interdisciplinary nature that couples unsteady aerodynamics, design, flight control and material selection \cite{festo2018,chen2025flapping}. Quadrotors provide precise hovering and aggressive maneuvering in clutter with comparatively simple mechanics, enabling inspection, cinematography, indoor logistics, and first response \cite{mahony2012multirotor,siciliano2016springer}. High control authority and a well-characterized mapping from motor commands to body forces and moments explain their agility under disturbance~\cite{7140074}. The mapping depends little on forward airspeed, and propeller thrust is well approximated by a quadratic function of angular velocity. Recent work demonstrates robust autonomous flight in challenging conditions \cite{10611562,doi:10.1126/scirobotics.abm6597}. In contrast, fixed-wing vehicles offer long endurance, high cruise efficiency, and wide-area coverage, which is well-suited to explore uncharted territory, environmental monitoring, and search mission (Fig.~\ref{fig:intro}A-C), but at the cost of limited low-speed agility and control effectiveness due to highly-coupled dynamics between flight parameters, such as airspeed, angle of attack and sideslip angles, just to name a few \cite{beard2012small}. 
In addition, drones in general need to consider a good trade-off between sensing and payload carrying capability, and while optical and inertial sensors are essential, they may not be able to sense clear-air turbulence, which is detrimental and poses life-threatening danger to the safe operation of any drones. To make matters worse, drones often maneuver at a fast rate, and thus, processing high-dimensional data like images may introduce significant delay in the control scheme.

To combine fixed-wing endurance with multirotor agility, we aim to learn a trustworthy mapping from control-surface deflections to aerodynamic forces and moments, supported by low-dimensional onboard sensors that reveal flow information not otherwise available. Achieving this goal requires two complementary capabilities: first, reliable real-time sensing of the local flow state; and second, models that capture nonlinear, sensor-dependent aerodynamics well enough to enable agile flight. To this end, we build an end-to-end sensing-to-control pipeline on a standard fixed-wing platform using \emph{sparse wing pressure sensors} and a \emph{low-cost, in-house multi-hole probe} (we use multi-hole probe, Pitot tube, and Pitot probe interchangeably). A central challenge is that near-body flow distortions induced by the fuselage, propeller slipstream, and control-surface deflections can contaminate pressure measurements. Inspired by standoff sensing in nature, such as the narwhal whale's tusk \cite{nweeia2014}, we mount an extended Pitot probe on a long boom (Fig.~\ref{fig:intro}D) to increase the sensor distance from the airframe, reduce self-induced aerodynamic interference, and promote earlier detection of incoming flow changes. We then introduce a three-part physics-informed, data-driven approach. First, we calibrate the multi-hole pressure measurements to estimate airspeed and flow angles. Second, we learn an observation-to-wrench model that links onboard sensing to aerodynamic forces and moments while enforcing a soft left--right symmetry prior. Third, we allocate control surfaces to track desired forces and moments using smooth, well-trimmed commands. Wind-tunnel experiments spanning a wide range of airspeeds, angles of attack, sideslip, and gust profiles show more accurate force estimation and improved tracking than unstructured baselines, with clear generalization beyond the training domain.

\section{Related Works}

\textbf{Fixed-Wing Drone Aerodynamics: }Fixed-wing drone aerodynamics presents several interrelated challenges. Unlike its bigger airliner counterpart that carries passengers, small drones have limited payload, rendering redundant sensing unrealistic, and its lightweight is often much more susceptible to atmospheric gust disturbance. At moderate to high angles of attack, airflow over the wing can separate, leading to abrupt loss of lift and unstable pitching behavior. Embedding pressure sensors within the wing or fuselage can provide early warning of flow separation, enabling corrective control \cite{Wood2019}. However, the wing itself induces an upstream updraft \cite{anderson2016}, so any pressure-based avionics must be carefully calibrated to account for this local flow distortion. Furthermore, changes in pitch (angle of attack) and yaw (sideslip) are coupled: when a fixed-wing drone banks, it not only needs differential flap and rudder adjustments to coordinate the turn, but also must increase airspeed to offset reduction of lift as part of it now becomes centripetal force \cite{handbook2025airplane}. These coupled effects make precise sensing and adaptive control essential for reliable maneuvering. Classic flight-control–oriented aerodynamic analysis \cite{nelson1998} relies heavily on aerodynamic derivatives—partial derivatives of aerodynamic forces and moments with respect to states such as angle of attack, sideslip angle, angular rates. This approach is valid primarily in small-angle regimes: the linearization assumes nearly steady, attached flow where lift and moment coefficients vary smoothly and predictably. 

\textbf{Fixed-Wing Drone Automation: }Research on fixed-wing drone automation is converging on three active directions. First, wind-aware guidance and robust control incorporate realistic gust models and disturbance observers to improve path following and disturbance rejection \cite{drones7040253}. Second, real-world reinforcement learning has progressed from simulation to outdoor flights by using domain randomization and latency/noise modeling, achieving data-efficient attitude control \cite{10101867}. Third, comparative evaluations under turbulence indicate that model-based approaches often match or exceed model-free methods and PID in tracking and robustness while yielding smoother actuation, although gains can shrink in severe perturbations \cite{olivares2024modelfreeversusmodelbasedreinforcement}. Recent surveys catalog tasks and toolchains and highlight the absence of standardized metrics and benchmarks, which limits fair comparison and reproducibility \cite{10609369}. These trends motivate integrated, wind-aware pipelines that combine sensing, structure-aware modeling, and control allocation for agile fixed-wing autonomy.

\textbf{Physics-Informed Machine Learning: } Research at the interface of classical modeling and learning has shown that embedding physical structure in learned models improves accuracy, interpretability, and sample efficiency \cite{10706036,doi:10.1126/scirobotics.abm6597,10155901,Djeumou2021NeuralNW}. Representative approaches encode governing laws or invariants directly in the architecture or loss, including PDE-constrained learning and PINNs \cite{raissi2019physics,NEURIPS2019_26cd8eca}, mechanics-aware networks that preserve energy or momentum \cite{cranmer2020lagrangianneuralnetworks,10598388}, and structure-aware representations that exploit symmetry and locality \cite{SanchezGonzalez2018GraphNA,xie2024morphological,11127302}. By constraining the hypothesis space to physically plausible behaviors, these methods improve robustness under distribution shift and reduce the reliance on dense instrumentation or exhaustive data collection. For fixed-wing flight, physics-informed learning is especially relevant because aerodynamic forces and moments depend on the airspeed in a strongly nonlinear way, and the mapping from sparse, partially observed sensors to these quantities is hard to identify. Instead of relying on PDE-constrained learning or PINNs, we encode domain structure directly by using how control surface deflections generate aerodynamics, and by utilizing fixed-wing drones left-right physical symmetries configuration, and baking it into the control-affine structure of the learned model.

\section{Hardware and System}
\begin{figure}
    \centering
    \includegraphics[width=1\linewidth]{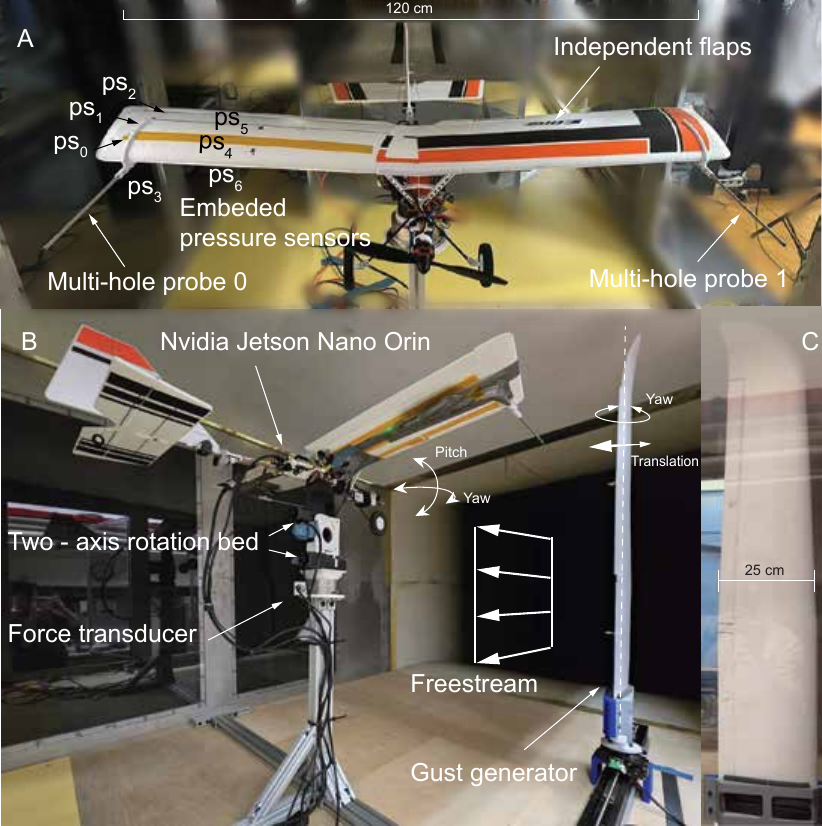}
    \caption{The model fixed-wing drone, model positioning system, and the gust generator in a wind tunnel. (A) The fixed-wing are instrumented with in-house developed avionics as well as an onboard Nvidia Jetson Nano Orin computer. (B) The test environment in the wind tunnel includes a gust generator as well as an in-house developed two-axis rotation bed (or model position system).}
    \label{fig:exprSetup}
\end{figure}

We design and construct a multi-hole probe system to measure local airflow direction and magnitude (Fig.~\ref{fig:exprSetup}A). The probe consists of five Albion aluminum microtubes ($0.5$ mm outer diameter (OD) × $0.3$ mm inner diameter (ID) arranged within a steel casing ($5$ mm OD × $4$ mm ID). Each aluminum tube is cut to approximately 32 cm in length and threaded through a black v4 resin-printed nose cone and end caps. Loctite super glue is applied to fix the tubes at both the nose cone and rear end caps. To facilitate connections with the onboard pressure sensors, each aluminum tube is linked to a short segment of silicon tubing ($3$mm OD x $1.5$ mm ID) through resin-printed connectors, which are tightly sealed with glue to prevent air leakage. These five tubes then attached directly to the bottom ports of 5 pressure sensors on a custom circuit board, which interface with a multiplexer routed to a Teensy 4.1 microcontroller, a compact ARM-based board with high-rate PWM/ADC and low-latency I/O for real-time interfacing. 

The model airplane (Fig.~\ref{fig:exprSetup}A) is a commercially available E-flite Slow Ultra Stick with a $1.2,\mathrm{m}$ wingspan. It uses a propeller, rudder, elevator, and left and right flaperons, which are combined flap–aileron surfaces driven independently. The spanwise size of the wind tunnel is $1.8$m, considerably larger than the drone, thus we are clear of any close-to-wall effect. We mount two in-house fabricated and calibrated five-hole probes close to the left and right wing tips, respectively, and they protrude about $30$~cm from the leading edge of the wing. Seven additional pressure sensor taps (named $ps_i$, where $i= 0$ to $6$) are placed on the right wing, flush with the surface. $ps_0$ - $ps_3$ are placed near the tip of the wing, and $ps_4$ - $ps_6$ are placed in the middle-span. In addition, $ps_0$ and $ps_4$ reside on the upper (suction side) of the leading edge, $ps_1$ and $ps_5$ mid-chord, and $ps_2$ trailing edge. $ps_3$ and $ps_6$ are also close to the leading edge, but are under the wing (pressure side). 

The model fixed-wing is then (Fig.~\ref{fig:exprSetup}B) placed on a two-axis in-house developed model positioning system, which varys the pitch (angle of attack) and yaw (sideslip angle) of the aircraft. An ATI six-axis force transducer records in realtime the forces and torques. The gust generator is a vertically-mounted wing, with length $1.2$m and chord $25$cm, and is positioned on the right side of the drone, $70$cm away from the center of the station in both the streamwise and spanwise direction. The gust generator is capable of yawing around its axis as well as translating back and forth on a linear stage. With a chord on the same order as the drone, the gust generator induces shear flow at small yaw angles and transitions to periodic vortex shedding as the yaw angle increases. The entire system is installed in a wind tunnel that produces clean, cyclic airflow with controllable airspeed.

For computing resources, an NVIDIA Jetson Orin Nano, a compact edge-AI module with an ARM CPU and a CUDA-capable GPU designed for embedded robotics, is mounted onboard and serves as the primary onboard computer. The Jetson sends control surface commands and receives pressure sensors measurements via the Teensy 4.1 microcontroller. The setup uses a workstation connected to the two-axis rotation rig, the force/torque transducer, and the gust generator. Both the workstation and the NVIDIA Jetson Nano run Ubuntu 20.04. The Jetson acts as the client, and the workstation acts as the system server. All actuation and inter-process communication use the robotic operating system (ROS) Noetic.  Clocks across devices are synchronized to ensure consistent timestamps and coordinated execution.

\section{Methodology}
\begin{figure*}[h]
    \centering
    \includegraphics[width=1.0\linewidth]{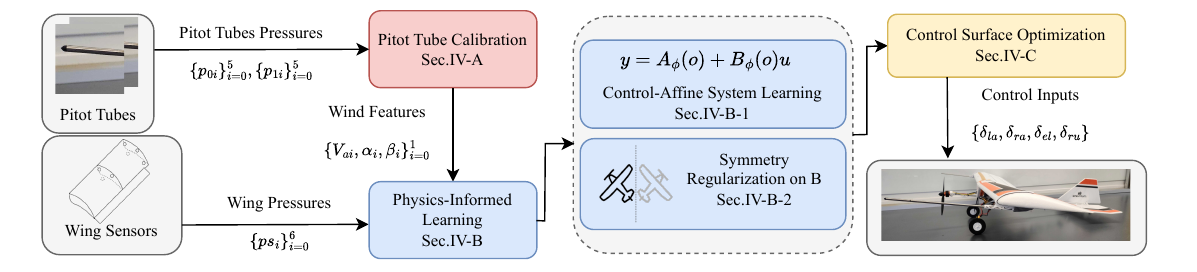}
    \caption{\textbf{Sensor-to-control pipeline.} \textbf{Section~\ref{subsec:alg_pitot}:} Dual multi-hole Pitot and sparse wing-pressure measurements are calibrated to estimate the flow state \((V_a,\alpha,\beta)\). \textbf{Section~\ref{subsec:alg_learning}:} A symmetry-aware, control-affine model maps onboard observations to aerodynamic forces and moments. \textbf{Section~\ref{subsec:opt}:} A convex controller then computes surface deflections to realize a desired wrench.}
    \label{fig:alg_chart}
    \end{figure*}
We present a sensing-to-control pipeline for small fixed-wing aircraft that couples
data-driven probe calibration, physics-informed dynamics learning, and convex control allocation, as summarized by the flow chart in Fig.~\ref{fig:alg_chart}. First, we calibrate the five-hole probes: from normalized pressures, we estimate a dynamic pressure correction and the flow angles, which enable robust recovery of wind features across different environments. Next, we learn a physics-informed, control-affine dynamics model that maps observations to forces and torques. Finally, we compute control-surface deflections by solving a small regularized least-squares problem that tracks a desired wrench.

\subsection{Pitot Tube Calibration}\label{subsec:alg_pitot}

The multi-hole probe is mounted on the airplane for \textit{in vivo} calibration, as the presence of the wing itself introduces updraft and would be better to consider these effects directly. 

We adopt a data-driven calibration for multi-hole velocity probes\cite{BirchDM2014Cauo, 8968380}. A probe tip with $n$ taps provides dimensional pressures $\{p_i\}_{i=1}^n$. Let
\[
p_{\max}=\max_i p_i,\qquad p_{\min}=\min_i p_i,\qquad \Delta p=p_{\max}-p_{\min},
\]
and define the nondimensional coefficients
\begin{equation}
C_{p_i}=(p_{\max}-p_i)\,/\Delta p,\qquad i=1,\ldots,n .
\label{eq:Cpi_assign}
\end{equation}
These coefficients are largely independent of airspeed and vary primarily with flow direction \cite{8968380}. The approximation error in using $\Delta p$ as a surrogate for dynamic pressure is captured by
\begin{equation}
C_d=\tfrac{1}{2}\rho V_a^2\,/\Delta p,
\label{eq:Cd_assign}
\end{equation}
where $\rho$ denotes air density and $V_a$ denotes the airspeed.

We realize the calibration with a neural network $f$ that consumes all normalized pressures and outputs the dynamic-pressure correction and flow angles explicitly:
\begin{equation}
C_d,\ \alpha,\ \beta \;=\; f\big(C_{p_1},\ldots,C_{p_n}\big),
\label{eq:nn_explicit}
\end{equation}
where $\alpha$ denotes the angle of attack relative to the probe axis and $\beta$ denotes sideslip angle relative to the probe axis. In deployment, we first compute $C_{p_i}$ using \eqref{eq:Cpi_assign}, evaluate \eqref{eq:nn_explicit} to obtain $C_d,\ \alpha,\ \beta$, and then reconstruct the airspeed by inverting \eqref{eq:Cd_assign}:
\begin{equation}
V_a=\sqrt{2\,\Delta pC_d/\rho}.
\label{eq:Va_assign}
\end{equation}

In our experiments, we use five-hole probes, where $n=5$.

\subsection{Physics-Informed Learning}\label{subsec:alg_learning}

Physics-informed learning augments data with a lightweight physics-based structure to improve sample efficiency and generalization. Classical physics-informed approaches (such as PINNs) typically penalize the PDE and boundary to enforce the governing laws~\cite{raissi2019physics}. In our low-speed fixed-wing setting with gusts and partial observability, we instead encode structural~\cite{xie2024morphological} and aerodynamic priors as constraints and regularizers on the network, which mitigates overfitting and yields a model whose inputs and outputs are naturally compatible with the control of fixed-wing drones.

Fixed-wing dynamics are nonlinear and vary under different wind conditions. Training directly on limited data tends to overfit to the training dataset and environment. Wing pressure sensors improve measurement coverage by capturing local flow effects that front-mounted probes miss. However, their signals are strongly coupled with flap deflections, which can introduce causal confounding and unstable gradients if used naively. We address these issues with a control-affine architecture and a symmetry-based regularizer on the control matrix, which improves parameter identification and stabilizes learning while allowing bounded asymmetric effects.

\subsubsection{Control-Affine System Learning}
Let $o$ denote the observation vector that aggregates flow-state features from two calibrated five-hole Pitot probes and wing pressure sensors. Specifically,
\[
o
=
\big[
\hat V_{a,0},\ \hat\alpha_{0},\ \hat\beta_{0},\
\hat V_{a,1},\ \hat\alpha_{1},\ \hat\beta_{1},\
ps_{0},\ldots, ps_{6}
\big]^\top,
\]
where $(\hat V_{a,j},\hat\alpha_{j},\hat\beta_{j})$ are the estimated airspeed, angle of attack, and sideslip from probe $j\in\{0,1\}$, and $\{ps_i\}_{i=0}^{6}$ are the wing pressure sensors measurements. Let the control vector be
\[
u=\big[\delta_{la},\,\delta_{ra},\,\delta_{el},\,\delta_{ru}\big]^\top\in\mathbb R^{4},
\]
comprising left flaperons, right flaperons, elevator, and rudder deflections. We predict the six-dimensional wrench
$y=[F_x,F_y,F_z,T_x,T_y,T_z]^\top\in\mathbb R^{6}$ with a control-affine model
\begin{equation}
    y \;=\; A_\phi(o) \;+\; B_\phi(o)\,u,
\end{equation}
where $A_\phi(\cdot)$ and $B_\phi(\cdot)$ are neural networks that share the first $k{-}1$ layers of a depth-$k$ backbone with neural network parameter $\phi$.

The control-affine parameterization is informed by fixed-wing aerodynamics. For relatively small angles of attack that are away from stall conditions, the incremental forces and moments induced by small control-surface deflections are approximately linear with deflection angles, and their effectiveness varies with the airspeed~\cite{stevens2015aircraft, etkin1995dynamics}. The term $A_\phi(o)$ captures the baseline aerodynamics due to the current flow and configuration, while $B_\phi(o)$ captures the state-dependent control effectiveness, which changes with dynamic pressure reflected by wind features measured by Pitot tubes and local loading reflected in the wing pressure sensors. By conditioning both $A$ and $B$ on $o$, the model respects known near-linear control effects, allows effectiveness to vary across operating points, and improves generalization relative to an unstructured mapping $y=F_\phi(o,u)$.

\subsubsection{Symmetry Regularization on \texorpdfstring{$B_\phi$}{B}}
Building on the control-affine parameterization, reliably identifying the flaperon-related columns of \(B_\phi(o)\) is challenging when training on datasets that combine wing-pressure measured on wings with flaperon commands. Without an explicit structural prior, \(B_\phi(o)\) often fails to recover the expected small-deflection aerodynamic patterns, which biases control-effectiveness estimates and degrades closed-loop performance. The inherent left–right symmetry of the flaperons provides a principled calibration signal for these columns~\cite{xie2024morphological, apraez2025morphological}. At the same time, real-world flow asymmetries like sideslip angles and gusts can cause the estimator to assign state-dependent effects unevenly between the two inputs. This motivates a soft, symmetry-aware regularization rather than a hard equality constraint.

To address this, we impose a physics-informed symmetry prior on the flaperon columns while keeping it soft to accommodate real asymmetries. Columns $0$ and $1$ of $B_\phi(o)\in\mathbb{R}^{6\times 4}$ correspond to the left and right flaperons. Aerodynamic symmetry implies opposite effects on some channels and same-sense effects on others. We encode this with the sign vector $\mathbf s=(1,-1,1,-1,1,-1)^\top$ and penalize deviations from the expected left–right relationship. We use a Huber-penalized regularizer that encourages symmetry but permits bounded deviations arising from sideslip and the upstream gust generator, so the model is not over-constrained in inherently asymmetric conditions. The loss is defined as:

\begingroup
\setlength{\abovedisplayskip}{4pt}
\begin{equation}
\mathcal L_{\mathrm{sym}}
= \lambda \sum_{i=1}^{6}
\psi_{\delta_i}\!\left(\big(B_{\phi:,0} + \mathbf{s}\odot B_{\phi:,1}\big)_i\right).
\end{equation}
with per-channel Huber thresholds \(\delta_i\) chosen by the user. The Huber penalty is
\[
\psi_{\delta}(e)=
\begin{cases}
\frac12 e^2, & |e|\le \delta,\\[2pt]
\delta\left(|e|-\frac12 \delta\right), & |e|>\delta.
\end{cases}
\]
\endgroup
Here \(\lambda\) scales the regularization, and \(\{\delta_i\}\) set the allowable deviation range for symmetry. This soft symmetry prior improves the practical identifiability of the two flaperon columns and limits leakage of wing-sensor correlations into control-effectiveness terms, while remaining faithful to conditions where symmetry does not hold.

\subsection{Control Surface Optimization}\label{subsec:opt}

We compute the control-surface deflections that realize a desired wrench at each
discrete time step $t\in\mathbb{N}$. Let $o_t$ denote the observation vector at
time $t$, $u_t\in\mathbb{R}^{4}$ the control input (left flaperon, right flaperon,
elevator, rudder), and $y_t\in\mathbb{R}^{6}$ the desired wrench
$[F_x,F_y,F_z,T_x,T_y,T_z]^\top$. Given the control–affine predictor
$y_t \approx A_\phi(o_t)+B_\phi(o_t)u_t$, we select $u_t$ by solving a convex, regularized
least-squares problem that balances tracking, smoothness, and trim adherence. For
compactness, we write $A_t\!\coloneqq\!A_\phi(o_t)$ and $B_t\!\coloneqq\!B_\phi(o_t)$. Let
$I$ be the $4\times4$ identity; $u_{t-1}$ the previous command; and $u_0$ a
fixed actuator-neutral trim. With nonnegative weights
$\lambda_0,\lambda_1\!\ge\!0$, we solve
\begin{equation}
\label{eq:rsls_obj}
\min_{u_t}\;
\underbrace{\|y_t-A_t-B_tu_t\|_2^2}_{\text{force tracking}}
\;+\;
\underbrace{\lambda_1\|u_t-u_{t-1}\|_2^2}_{\text{smooth regularization}}
\;+\;
\underbrace{\lambda_0\|u_t-u_0\|_2^2}_{\text{damping term}}.
\end{equation}

\emph{Force tracking} enforces consistency between the predicted wrench
$A_t+B_tu_t$ and the target $y_t$. \emph{Smooth regularization} penalizes
departures from $u_{t-1}$ to promote temporal smoothness and avoid abrupt
actuator changes. \emph{Damping} biases the command toward the neutral trim
$u_0$, reducing magnitude, helping prevent saturation, and improving numerical
conditioning. Problem \eqref{eq:rsls_obj} is quadratic and strictly convex when
$\lambda_0+\lambda_1>0$, thus admitting a closed form. Setting the gradient to
zero yields
\begin{equation}
\big(B_t^\top B_t + (\lambda_0+\lambda_1)I\big)u_t
=
B_t^\top(y_t-A_t) + \lambda_1 u_{t-1} + \lambda_0 u_0.
\end{equation}
Define
\[
Q_t = 2\big(B_t^\top B_t + (\lambda_0+\lambda_1)I\big),
\]
\[
c_t = 2\big(B_t^\top(y_t-A_t) + \lambda_1 u_{t-1} + \lambda_0 u_0\big),
\]
and compute the optimal command as the unique solution to \(Q_t u_t^\star = c_t\).
Positive $(\lambda_0,\lambda_1)$ render $Q_t$ positive definite and trade off tracking
fidelity against command smoothness and damping, typically yielding more stable,
well-conditioned inputs in closed-loop operation.

\section{Experiments}

\begin{figure}[h]
    \centering
    \includegraphics[width=1.0\linewidth]{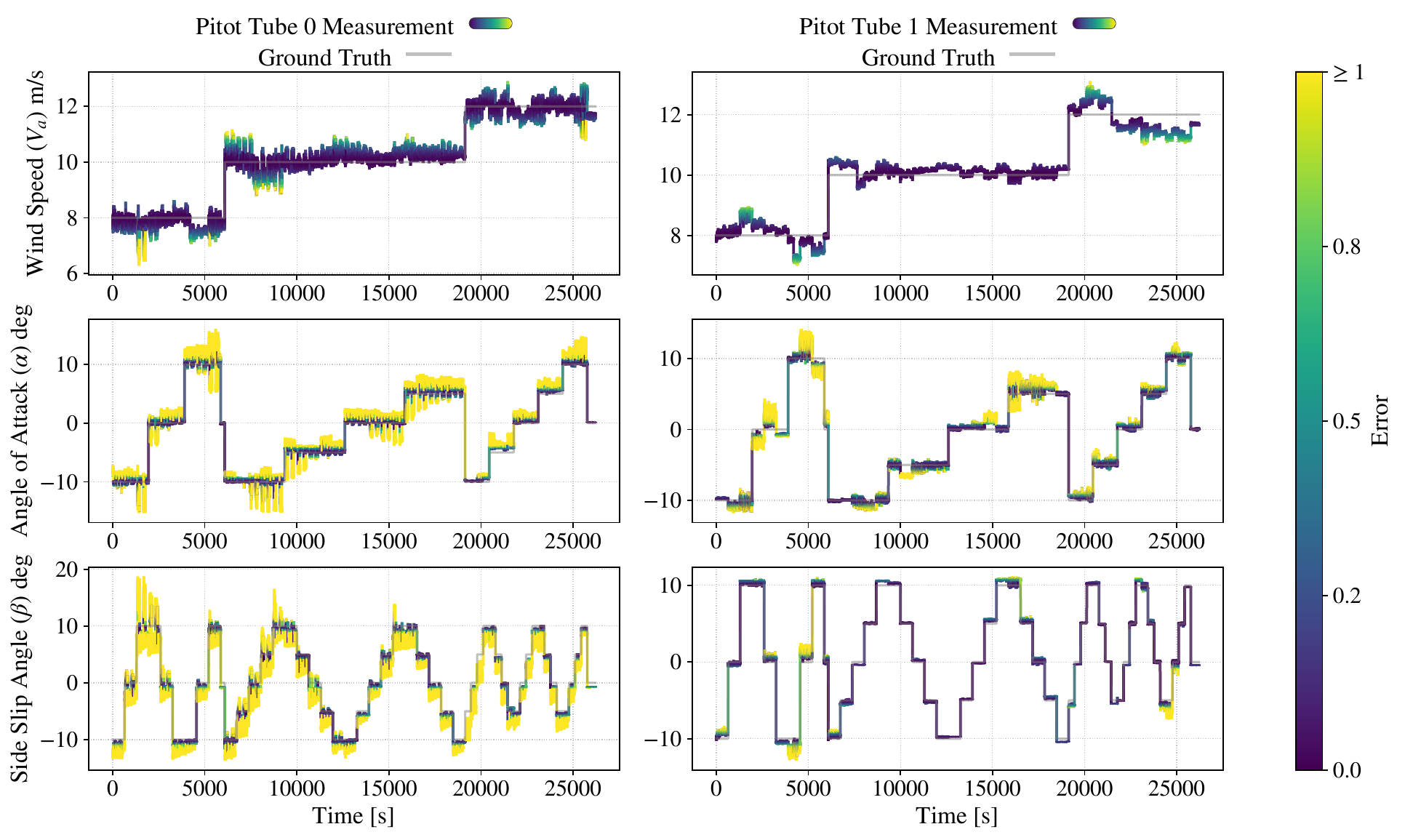}
    \caption{\textbf{Estimated vs. ground-truth wind features.} Columns: Pitot tube 0 (left) and Pitot tube 1 (right). Rows: airspeed \(V_a\), angle of attack \(\alpha\), and sideslip angle \(\beta\). The gust is positioned near probe 0.}
    \label{fig:pitot_calibration}
\end{figure}

We conduct all experiments in a wind tunnel with a cross section of $1.3$~m tall and $1.8$~m wide. All learning components are implemented in Python using the PyTorch library~\cite{paszke2019pytorch}. We trained and validated the Pitot tube calibration model on a grid of airspeeds \(V_a \in \{8,10,12\}\,\mathrm{m/s}\), angles of attack \(\alpha \in \{-10^\circ,-5^\circ,0^\circ,5^\circ,10^\circ\}\), and sideslip angles \(\beta \in \{-10^\circ,-5^\circ,0^\circ,5^\circ,10^\circ\}\). For validation, we additionally actuated the upstream gust generator to introduce unsteady disturbances. The results are shown in Fig.~\ref{fig:pitot_calibration}. Both probes recover the flow features \((V_a,\alpha,\beta)\) with good accuracy, though errors remain; in particular, probe~0 exhibits noticeably larger errors when the gust generator is active and positioned close to the test article. This highlights the fact our gust generator produces considerable amount of turbulence to our model airplane. 

Next, we train force models on the collected dataset and evaluate them using baselines and ablations. Our proposed model, \textit{Affine Sym}, is a control-affine network \(y=A_\phi(o)+B_\phi(o)u\) with a soft left–right symmetry regularizer on the flaperon columns of \(B_\phi(o)\). The variant \textit{Affine} removes the symmetry regularization while retaining the control-affine structure. To assess the contribution of wing-sensor inputs, \textit{Affine w/o WS} excludes wing-pressure measurements from the observation vector. As non-affine baselines, \textit{Unstructured} denotes a standard MLP mapping \((o,u)\mapsto y\), and \textit{Unstructured w/o WS} further removes wing pressure inputs. Throughout, “WS” refers to wing sensor pressures.

We first assess the contribution of wing–pressure sensors to forces and moments learning. 
At \(V_a=10~\mathrm{m/s}\), Table~\ref{tab:ws} reports root–mean–square errors (RMSE) for the six-dimensional wrench—forces (N) and moments (\(\mathrm{N\cdot m}\)). 
Due to the page limit, we report a single RMSE over the concatenated 6D wrench (forces in N, torques in N·m) for all tables in this paper. This mixes units and should be read as a comparative aggregate.
Incorporating wing-pressure measurements into the network’s input reduces error by \(24.6\%\)–\(29.5\%\) relative to the corresponding models without wing-sensor inputs. 
Based on experimental observations, the wing-mounted pressure sensors are located close to the upstream gust generator. Their readings, particularly from the leading-edge sensors \(ps_0\) and \(ps_4\), vary systematically with the gust angle. Including these wing-pressure signals as inputs therefore provides local context about gust-induced unsteady flow, improving calibration and estimation under disturbed conditions.

\begin{table}[h]
\caption{Estimation Errors \((N/N\cdot m)\) under $10 \ m/s$ Wind}
\label{tab:ws}
\begin{center}
\begin{tabular}{|c|c|c|c|}
\hline
Affine & Affine w/o WS & Unstructured & Unstructured w/o WS\\
\hline
$0.450$ & $0.597$  & $\mathbf{0.442}$ & $0.627$ \\
\hline
\end{tabular}
\end{center}
\end{table}

\begin{figure}[h]
    \centering
    \includegraphics[width=1.0\linewidth]{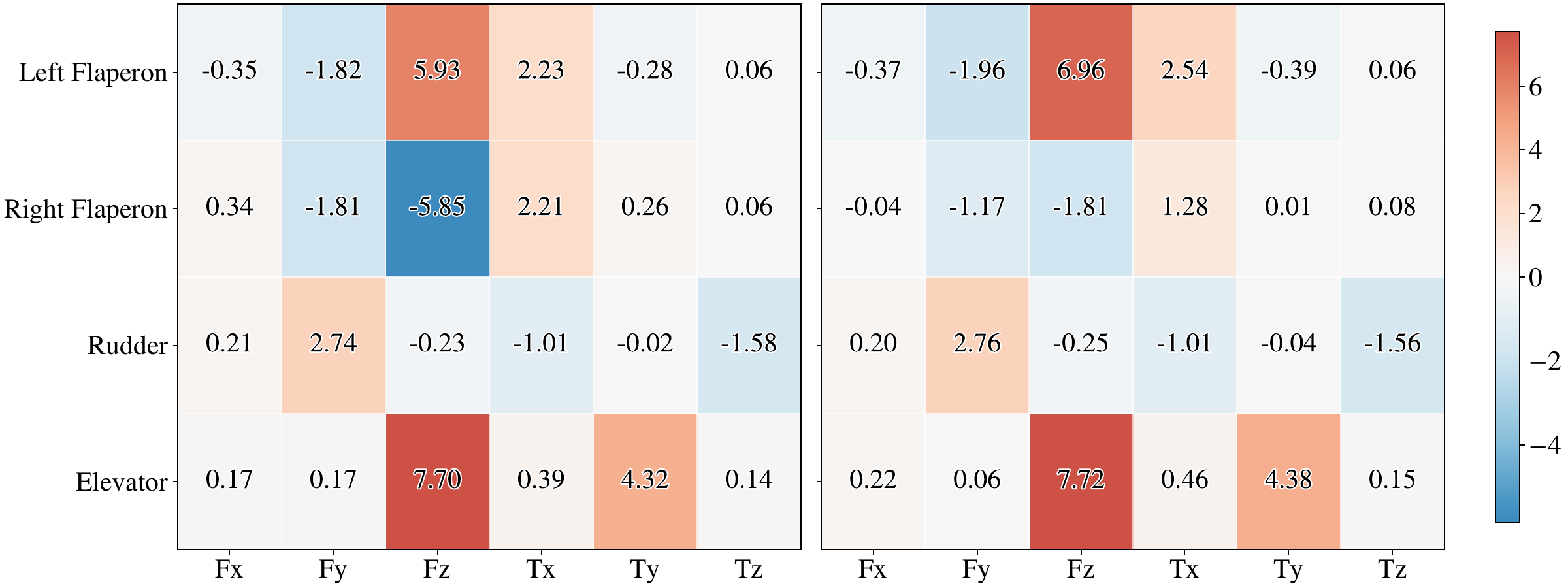}
    \caption{\textbf{\(B_\phi^\top\) heatmaps.} Left: model trained with symmetry regularization; right: without symmetry regularization. The regularized map is more aerodynamically plausible.}
    \label{fig:B_heatmap}
\end{figure}

However, adding wing-pressure readings without additional structure introduces confounding. Because these signals respond directly to the right-flaperon deflection, supplying them unregularized creates strong endogeneity between the observation channels and the control input. The network can then explain flaperon-induced changes through the wing pressure inputs rather than the right-flaperon control input, so the estimated control-effectiveness matrix \(B_\phi(o)\) assigns insufficient sensitivity to the right-flaperon column and fails to capture the true control–force relationship. This misattribution is evident in the right plot of Fig.~\ref{fig:B_heatmap}, where the right-flaperon column appears inconsistent. As an illustration on the lift channel \(F_z\), the estimated control-effectiveness entries in \(B_\phi\) are \(B_{F_z,\delta_{ra}}=-1.81\) (right flaperon) and \(B_{F_z,\delta_{la}}=6.96\) (left flaperon), i.e., the left-flaperon magnitude is \(\approx 3.8\times\) that of the right, whereas in reality, the actuation of left and right flaperons should have the same aerodynamic impact on the lift force \(F_z\). This pronounced imbalance indicates that, without regularization, the model misattributes flaperon-induced lift changes, yielding a \(B_\phi\) matrix that does not faithfully represent the true input–output mapping.

To mitigate this issue, we impose a soft left–right symmetry regularization on \(B_\phi(o)\). The regularizer penalizes deviations of the left- and right-flaperon columns from the expected sign pattern implied by aerodynamic symmetry, while allowing bounded departures to accommodate genuine asymmetries induced by sideslip and the gust generator. As shown in the left panel of Fig.~\ref{fig:B_heatmap}, the resulting estimate of \(B_\phi\) restores a consistent right-flaperon sensitivity.

\begin{figure}[h]
    \centering
    \includegraphics[width=1.0\linewidth]{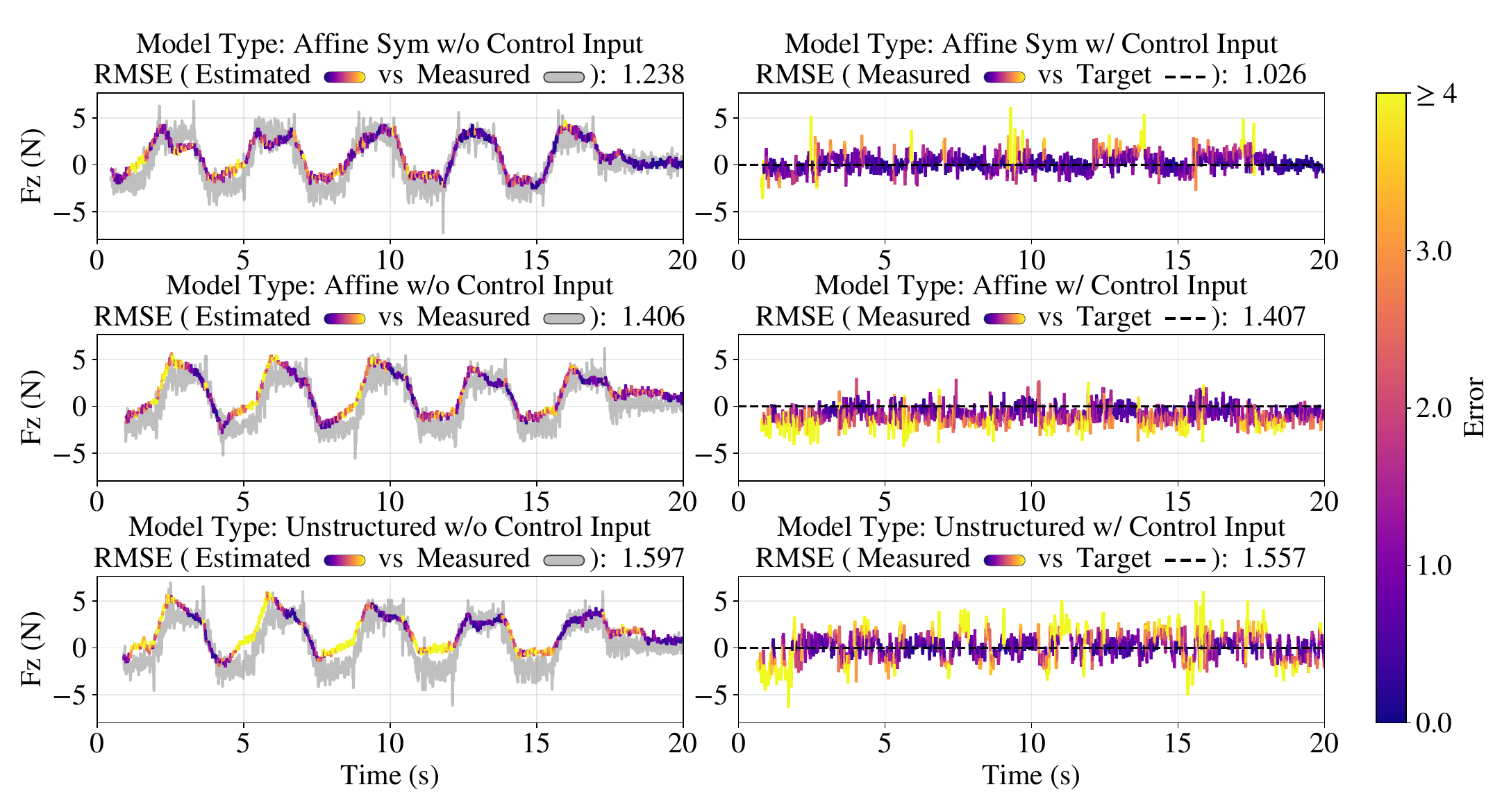}
    \caption{\textbf{Normal force \(F_z\) estimation under wind speed 14\(\mathbf{m/s}\).} Columns: left, Measured vs.\ Estimated without control input; right, Measured vs.\ Target with control input. Rows: \textit{Affine Sym}, \textit{Affine}, and \textit{Nonlinear} models. The results indicate that the \textit{Affine Sym} model yields the lowest estimation and tracking errors.} 
    \label{fig:fz_compare}
\end{figure}
\begin{table}[h]
\caption{Estimation Errors \((N/N\cdot m)\) Under Different Wind Speeds}
\label{tab:est_errors}
\begin{center}
\begin{tabular}{|c|c|c|c|}
\hline
Wind Speed & Affine Sym & Affine & Unstructured\\
\hline

$7$ $m/s$ & $\mathbf{0.521}$  & $0.531$ & $0.521$ \\
$9$ $m/s$ & $0.495$ & $0.493$ & $\mathbf{0.484}$ \\
\color{blue}{$10$} $m/s$ & \color{blue}{$0.522$} & \color{blue}{$0.450$} & \color{blue}{$\mathbf{0.442}$} \\
$14$ $m/s$ & $\mathbf{0.592}$ & $0.598$ & $0.637$ \\
\hline
\end{tabular}
\end{center}
\end{table}
We evaluated the trained force–prediction models under an in-distribution condition (\(V_a=10~\mathrm{m/s}\)) and under out-of-distribution wind speeds \(V_a\in\{7,9,14\}~\mathrm{m/s}\) to assess generalizability. Unlike the training settings, the validation experiments comprises two stages. 
\textbf{Stage I:} the angle of attack \(\alpha\), sideslip \(\beta\), and gust generator actuation are varied simultaneously and continuously to induce time-varying flow conditions. 
\textbf{Stage II:} \(\alpha\) and \(\beta\) are held at prescribed setpoints while the gust generator remains active, allowing us to check whether the aircraft can correctly estimate forces and torques and remain balanced under turbulence. These scenarios test whether the system can (i) accurately estimate the instantaneous wrench and (ii) generate control inputs that realize a specified target wrench under such conditions. Table~\ref{tab:est_errors} reports the RMSE between predicted forces and measurements from the force transducer. The \textit{Unstructured} baseline exhibits clear overfitting, with errors growing markedly at higher airspeed. Its RMSE increases by \(44.1\%\) from \(10\) to \(14~\mathrm{m/s}\). In contrast, \textit{Affine Sym} degrades more gracefully, with only an \(11.8\%\) increase over the same shift, indicating substantially better out-of-distribution robustness.
\begin{table}[h]
\caption{Tracking Errors \((N/ N\cdot m)\) with Control}
\label{tab:tracking_error}
\begin{center}
\begin{tabular}{|c|c|c|c|}
\hline
Wind Speed & Affine Sym & Affine & Unstructured\\
\hline
$7$ $m/s$ & $\mathbf{0.635\pm 0.13}$ & $0.678\pm 0.13$ & $0.748 \pm 0.14$ \\
$9$ $m/s$ & $0.660 \pm 0.06$ & $\mathbf{0.648\pm 0.09}$ & $0.791 \pm 0.01$ \\
\color{blue}{$10$} $m/s$ & \color{blue}{$0.609\pm0.07$} & \color{blue}{$\mathbf{0.598\pm 0.07}$} & \color{blue}{$0.603\pm0.07$} \\
$14$ $m/s$ & $\mathbf{1.287 \pm 0.07}$ & $1.309 \pm 0.07$ & $1.417 \pm 0.02$ \\
\hline
\end{tabular}
\end{center}
\end{table}
We visualize the time series of the estimated normal force \(F_z\) for Stage-I in the left column of Fig.~\ref{fig:fz_compare}, with the per-subplot RMSE reported above each panel. The \textit{Unstructured} baseline exhibits substantially larger errors than the \textit{Affine} and \textit{Affine Sym} models across conditions, consistent with overfitting. These results support the advantage of the control-affine formulation: by constraining the mapping to \(y=A_\phi(o)+B_\phi(o)u\), it generalizes more reliably in the limited-data scenarios.

\begin{table}[h]
\caption{RMSSD for Control Input under $14 \ m/s$ Wind}
\label{tab:rmssd}
\begin{center}
\begin{tabular}{|c|c|c|c|}
\hline
Control Input & Affine Sym & Affine & Unstructured\\
\hline
Right Flaperon & $\mathbf{16.30\pm1.65}$ & $39.27\pm 3.60$ & $57.78\pm30.35$ \\
Average & $\mathbf{13.46\pm 1.36}$ & $17.96\pm 2.40$ & $49.54\pm 31.10$ \\
\hline
\end{tabular}
\end{center}
\end{table}

Finally, we evaluate whether the learned model, combined with our control-surface optimizer, can track and maintain a desired aerodynamic force across varying wind conditions. In this experiment, the angle of attack, sideslip, and gust generator actuation vary simultaneously and continuously, while the optimizer computes control-surface deflections \(\{\delta_{la},\delta_{ra},\delta_{el},\delta_{ru}\}\) to realize a prescribed target wrench. Across all experiments, every model uses the same regularization weights \(\lambda_0\) and \(\lambda_1\). Table~\ref{tab:tracking_error} reports tracking RMSEs across three cases, namely extrapolated out-of-distribution airspeeds, an interpolated within-range airspeed at \(V_a = 9~\mathrm{m/s}\) that is held out, and an in-distribution airspeed at \(V_a = 10~\mathrm{m/s}\). In the extrapolated case, \textit{Affine Sym} attains the lowest error. At \(V_a = 9~\mathrm{m/s}\), \textit{Affine Sym} and \textit{Affine} both outperform the unstructured model and are comparable, with \textit{Affine} slightly lower. At \(V_a = 10~\mathrm{m/s}\), differences among all methods are small. 

We run three rounds of experiments at different airspeeds to guard against run-to-run variability and evaluate consistency. The analysis focuses on \(V_a=14~\mathrm{m/s}\), which lies in the extrapolative regime where generalization is most demanding. A representative Stage-I control sequence is shown in the right column of Fig.~\ref{fig:fz_compare}. In these plots, the \textit{Affine} model exhibits a small but persistent bias relative to the target, which is consistent with estimation error, while the \textit{Nonlinear} baseline shows noticeably less stable behavior. On the \(F_z\) channel over this interval, \textit{Affine Sym} reduces the tracking RMSE by \(27\%\) relative to \textit{Affine} and by \(34\%\) relative to \textit{Nonlinear}.

These results indicate that enforcing a control-affine structure improves generalization across conditions, while the symmetry prior further strengthens extrapolation beyond the training range. Although symmetry regularization improves identifiability, it is not universally optimal for fixed-wing systems. Aerodynamic and actuation asymmetries can arise from many reasons, so enforcing a strict mirror prior with a finite tolerance may mis-specify the model in some regions. This helps explain why \textit{Affine Sym} does not always have the top performance in certain interpolated cases, where the approximate prior can conflict with the true physics. Under extrapolation, however, the symmetry prior provides a stronger inductive bias and yields the best generalization among the compared models. Beyond accuracy, the symmetry regularization also promotes steadier control, as reflected by lower root mean square of successive differences (RMSSD) values in Table~\ref{tab:rmssd} at airspeed \(V_a = 14~\mathrm{m/s}\). The control penalties \(\lambda_0\) and \(\lambda_1\) introduce tolerance to inaccuracies in the \(B_\phi\) and enable force tracking where the estimated forces are reliable, but they can induce oscillations in the control inputs, which RMSSD captures quantitatively.

\section{Conclusion}

This paper presents a sensing-to-control pipeline for fixed-wing UAVs that combines bio-inspired hardware, physics-informed learning, and convex control allocation. A narwhal-inspired standoff multi-hole probe increases sensor standoff to reduce near-body flow distortion, while sparse wing pressure sensors provide local context under unsteady conditions. A data-driven calibration estimates airspeed and flow angles from normalized pressures, and a control-affine force model with a soft left--right symmetry regularizer improves identifiability of state-dependent control effectiveness under partial observability. Desired aerodynamic wrenches are tracked via a small regularized least-squares allocator.

Wind-tunnel experiments across multiple airspeeds and gust conditions show improved estimation, robustness, and actuation. Adding wing pressures reduces force-estimation error by \(25\%\) to \(30\%\). Under distribution shift, performance degrades by \(\sim 12\%\), compared with \(44\%\) for an unstructured baseline. In closed loop, normal-force tracking error drops by \(27\%\) versus a plain affine model and by \(34\%\) versus a nonlinear baseline, with smoother inputs, supporting the benefit of standoff sensing with physics-informed learning.

Future work will incorporate temporal dynamics and broaden operating regimes. Observation histories can help infer turbulence and vortices, estimate gust location/strength/angle, and forecast dynamics under inconsistent flow. We will also integrate uncertainty-aware free-flight control by embedding observation and environment uncertainty into an MPC framework for stability and safety, study stall in free flight, and refine the symmetry prior and sensor placement to improve robustness and generalization.

\section*{ACKNOWLEDGMENT}

We thank J. Humml for guidance on fabricating our pitot probes, R. Nemovi for help evaluating the Ultra Slow Stick as a payload-capable airframe, and M. Anderson for insightful input throughout the project.

\bibliographystyle{IEEEtran}  %
\bibliography{references}

\addtolength{\textheight}{-12cm}   %

\end{document}